\documentclass[letterpaper]{article} % DO NOT CHANGE THIS
\usepackage{aaai25}  % DO NOT CHANGE THIS
\usepackage{times}  % DO NOT CHANGE THIS
\usepackage{helvet}  % DO NOT CHANGE THIS
\usepackage{courier}  % DO NOT CHANGE THIS
\usepackage[hyphens]{url}  % DO NOT CHANGE THIS
\usepackage{graphicx} % DO NOT CHANGE THIS
\usepackage{natbib}  % DO NOT CHANGE THIS AND DO NOT ADD ANY OPTIONS TO IT
\usepackage{caption} % DO NOT CHANGE THIS AND DO NOT ADD ANY OPTIONS TO IT
\usepackage{algorithm}
\usepackage{algorithmic}
\usepackage{tabularray}
\usepackage{graphicx}
\usepackage{booktabs}
\usepackage{array}
\usepackage{enumitem}
\usepackage{multirow}
\usepackage{newfloat}
\usepackage{listings}
\usepackage{verbatim}
\DeclareCaptionStyle{ruled}{labelfont=normalfont,labelsep=colon,strut=off} % DO NOT CHANGE THIS
\floatstyle{ruled}
\newfloat{listing}{tb}{lst}{}
\floatname{listing}{Listing}
\nocopyright %-- Your paper will not be published if you use this command
\title{``My Grade is Wrong!'': A Contestable AI Framework for Interactive Feedback in Evaluating Student Essays}
\author{
    Shengxin Hong\textsuperscript{\rm 2},
    Chang Cai\textsuperscript{\rm 1},
    Sixuan Du\textsuperscript{\rm 3},
    Haiyue Feng\textsuperscript{\rm 4},
    Siyuan Liu\textsuperscript{\rm 1},
    Xiuyi Fan\textsuperscript{\rm 1}
}
\affiliations{
    \textsuperscript{\rm 1}Nanyang Technological University, Singapore\\
    \textsuperscript{\rm 2}Hubei University of Technology, China\\
    \textsuperscript{\rm 3}Hohai University, China\\
    \textsuperscript{\rm 4}Shenzhen University, China\\
    xyfan@ntu.edu.sg
}

\begin{document}

\maketitle

\begin{abstract}
Interactive feedback, where feedback flows in both directions between teacher and student, is more effective than traditional one-way feedback. However, it is often too time-consuming for widespread use in educational practice. While Large Language Models (LLMs) have potential for automating feedback, they struggle with reasoning and interaction in an interactive setting. This paper introduces CAELF, a Contestable AI Empowered LLM Framework for automating interactive feedback. CAELF allows students to query, challenge, and clarify their feedback by integrating a multi-agent system with computational argumentation. Essays are first assessed by multiple Teaching-Assistant Agents (TA Agents), and then a Teacher Agent aggregates the evaluations through formal reasoning to generate feedback and grades. Students can further engage with the feedback to refine their understanding. A case study on 500 critical thinking essays with user studies demonstrates that CAELF significantly improves interactive feedback, enhancing the reasoning and interaction capabilities of LLMs. This approach offers a promising solution to overcoming the time and resource barriers that have limited the adoption of interactive feedback in educational settings.
\end{abstract}

\section{Introduction}
\begin{figure}[!htbp]
	\centering
	\includegraphics[width=1\columnwidth]{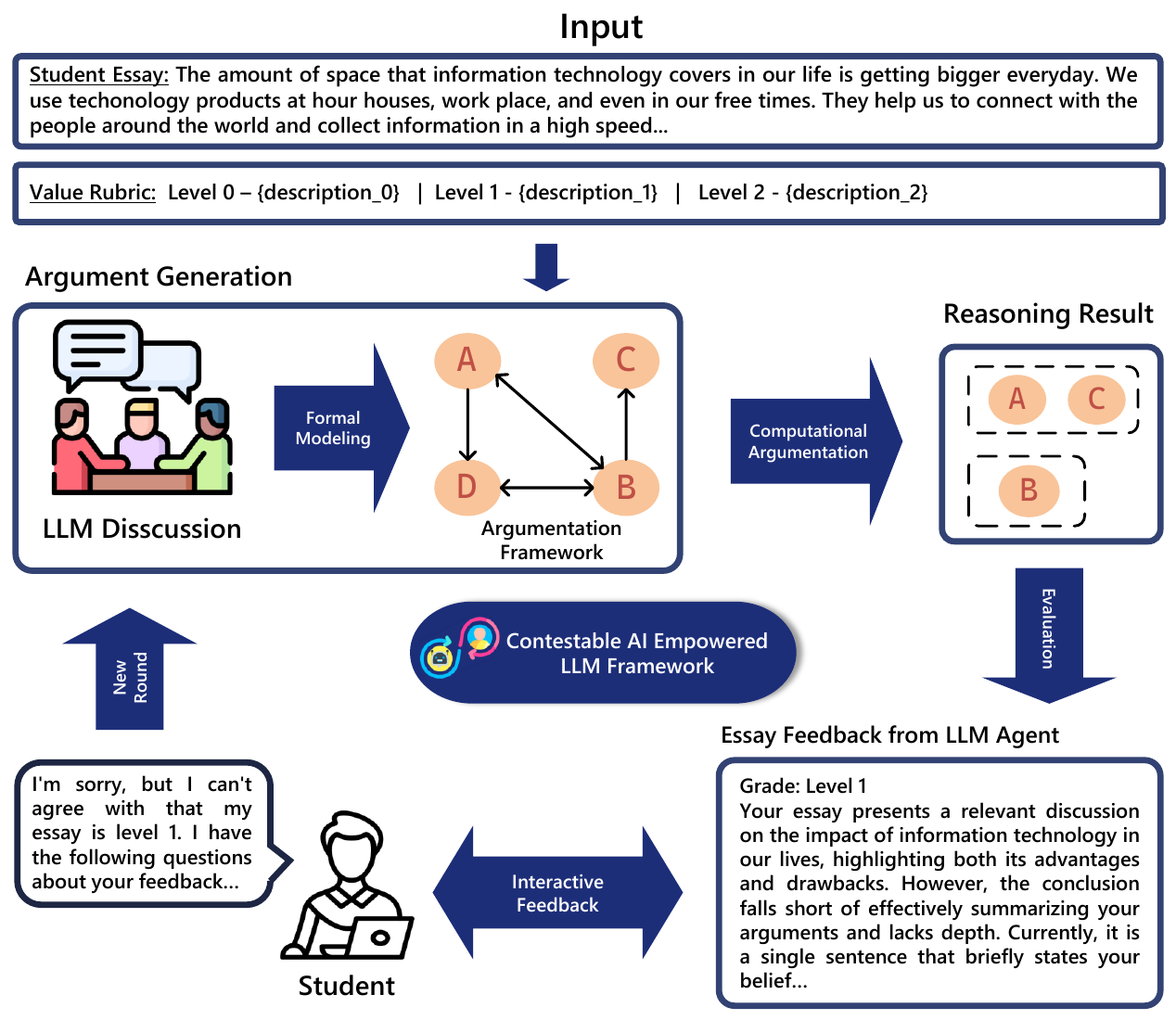}
	\caption{Diagram of our contestable AI empowered LLM framework for interactive feedback generation (CAELF).} 
	\label{fig}
\end{figure}
As stated in Hattie and Timperley’s landmark paper \cite{doi:10.3102/003465430298487}, 
\begin{quote}
{\em Feedback is one of the most powerful influences on learning and achievement},     
\end{quote}
the question of how best to provide effective feedback to students has been a long-standing research question in education. For instance, \cite{sadler1989formative} has emphasized the importance of formative assessment and its role in helping students understand the standards they are aiming for, while \cite{shute2008focus} has explored the idea of formative feedback that is timely, specific, and focused on the learning process.
%, arguing that such feedback can significantly enhance student achievement. 

More recently, Nicol \cite{nicol2014monologue} has proposed considering {\bf Interactive Feedback} as an alternative feedback format. Nicol suggests that feedback should not merely be a one-way transmission of information from instructor to student. Instead, it should involve students actively engaging with the feedback, interpreting it, and using it as a basis for further learning and improvement. However, despite the benefits of interactive feedback, it is not commonly used in practice, as it is time-consuming for teachers and requires significantly more investment in preparation and teaching delivery \cite{hopfenbeck2023challenges,westera2018automated}.

Large Language Models (LLMs) have demonstrated success in applications such as automatic scoring and feedback generation \cite{dai2023can,gubelmann2024exploring,kostic2024llms}. For example, \cite{10260740} highlights how LLMs enable educators to provide feedback to larger classes more efficiently. As a result, LLM-empowered interactive feedback has emerged as a promising approach to overcome the time and resource constraints that have historically hindered the widespread adoption of interactive feedback. However, a review of past efforts to explore LLM capabilities \cite{moore2023assessing,wang2023chatgptdefendbelieftruth,xiu-etal-2022-logicnmr} reveals that existing LLM techniques, such as prompt engineering, fall short in meeting two critical requirements for effective interactive feedback:
\begin{enumerate} 
\item Reasoning: Interactive feedback is formative, requiring the continuous reception of information from students and the generation of targeted responses to guide their learning. LLMs, however, may struggle with accurately understanding and reasoning through subtle nuances in each round of interaction, potentially leading to incorrect assessments and incomplete feedback.

\item Interaction: LLMs are prone to issues of over-reliance on user input, which undermines their ability to provide reliable interactive feedback. While LLMs may initially generate correct responses, they often falter when challenged by illogical or invalid arguments from users, failing to maintain consistency in their feedback. 
\end{enumerate}

In interactive feedback, student responses may be partially incorrect, and vocabulary and stylistic expressions may vary widely \cite{liu2016validation}, which can lead to incorrect grading and incomplete feedback from the LLM after the interaction. Consider the example illustrated in Table~\ref{tabel:exp}. In this scenario, an LLM is tasked with providing feedback on an essay about critical thinking. Although the initial feedback demonstrates valid reasoning, when the user challenges it, the LLM mistakenly retracts its conclusion and concedes to the user’s invalid argument.

\begin{table}
\begin{small}
	\centering
	\begin{tabular}{p{0.96\linewidth}}
		\toprule
		\textbf{LLM:} Your essay effectively argues that online criminal activities justify controlling internet use for security reasons, while also acknowledging the efficiency benefits of unrestricted online activities. By presenting both a positive and a negative argument on the issue, your essay meets the \emph{Level 1} criteria.  \\\\
		\textbf{Student:} I believe the rubric is wrong, so my grading is wrong. With one positive and one negative argument, my grading should be {\em level 2}. My arguments on preventing criminal activities and ensuring safety by screening dangerous content are both very strong. Thus, {\em level 2} must be my grading. \\\\
		\textbf{LLM:} Based on your response, your essay should be graded at \emph{Level 2}. You provided two strong arguments for controlling internet use and acknowledged the opposing view by noting how unlimited online activities can improve work efficiency. This balance meets the criteria for {\em Level 2}. \\		
		\bottomrule
	\end{tabular}
\caption{An example where an LLM provided correct feedback and grading, but when confronted with the student's challenge, it incorrectly altered the grade and feedback.
\label{tabel:exp}}
\end{small}
\end{table}

%Although some studies have considered the human-in-the-loop LLM for automated assessment \cite{cohn2024chain,cohn2024towards}, they have mainly focussed on specific domain or scenario and have not systematically addressed the challenges that exist in the application of LLM techniques to interactive feedback.

%To address this gap, 
In this work, we propose a {\bf Contestable AI Empowered LLM Framework for Interactive Feedback Generation (CAELF)} as illustrated in Figure~\ref{fig}. CAELF takes students' essays and an assessment rubric as its inputs and produces feedback as its outputs. CAELF supports interactive feedback as it allows students to enquiry and challenge its feedback and provide additional justifications as needed. 

%To address the identified weaknesses of existing LLMs, 
CAELF employs a Contestable AI paradigm based on a multi-agent argumentation system. At a high level, each essay is first examined independently by several ``Teaching-Assistant Agents (TA agents).'' Each TA agent represents a specific aspect of the essay evaluation as outlined by the assessment rubrics. Arguments are then formed by aggregating the evaluations made by TA agents, initiating a formal argumentation process to determine the essay grade and generate summary feedback. Subsequently, users have the opportunity to challenge the argumentation process for further clarification. As shown in our experiments, CAELF with its formal reasoning addresses both weakness of existing LLMs. 

To evaluate the effectiveness of CAELF, we conducted a case study on the evaluation of critical thinking essays. Using a dataset of 500 essays and a four-dimensional evaluation rubric (``issue,'' ``evidence,'' ``position,'' ``conclusion''), CAELF demonstrated (1) initial grading accuracy comparable to GPT-4 across all four dimensions, and (2) significantly better performance in interaction grading accuracy and maintaining consistent evaluations despite user challenges. Additionally, in a separate human user study, we found that CAELF's feedback outperformed the baselines in terms of ``factual accuracy,'' ``self-regulation,'' and ``suggestions for future improvement.''

Our contributions are summarized as follows:
\begin{enumerate}
    \item CAELF is the first framework to integrate contestable AI design into LLM feedback generation within the educational domain, addressing LLM shortcomings in providing interactive feedback.
    
    \item CAELF utilizes argumentation frameworks for formal reasoning, enhancing explainability and allowing decisions made by the LLM to be challenged by humans.
    
    \item CAELF fosters reflective learning environments through agent collaborative discussions (human-LLM or LLM-LLM), encouraging students to deepen their understanding. Experimental results on a dataset of 500 critical thinking essays, along with extensive human evaluation, jointly demonstrate CAELF's effectiveness.
\end{enumerate}

\section{Related Work and Background}
\subsubsection{LLMs for Essay Evaluation and Feedback}

LLMs have become increasingly popular in automating essay evaluation and feedback generation, reducing the manual effort traditionally required \cite{kostic2024llms}. They have shown promise in automating scoring, cutting down on time and labor \cite{boud2013rethinking,dai2023can}. For example, \citeauthor{yancey2023rating} demonstrated that GPT-4 can evaluate short English essays with near-equal performance to modern Automatic Writing Evaluation (AWE) methods, without specific training. Additionally, LLMs also can generate clear, natural language feedback that explains the reasoning behind them, enhancing transparency in the evaluation process \cite{dai2023can, cohn2024chain}. This capability is particularly valuable in educational settings, as it helps bridge the gap between evaluation and learning. Studies show that students find LLM-generated feedback helpful and rate its quality as good to very good \cite{gubelmann2024exploring}. However, challenges persist in accurately grading complex texts, fine-tuning, and providing tailored feedback. \cite{kostic2024llms} highlighted LLM limitations in evaluating complex academic texts, showing a gap between LLM capabilities and the nuanced requirements of student essay evaluation. Moreover, \cite{stahl2024exploring} found that LLM-generated feedback does not sufficiently leverage specific scores to enhance its relevance and actionability.

\subsubsection{Contestable AI}
Contestable AI asserts that models used in critical tasks like decision-making or evaluation should enable users to question, contest, and review their outputs \cite{alfrink2023contestable,10.1145/3064663.3064703}. \cite{leofante2024contestableaineedscomputational} argue that contestable AI requires computational argumentation, allowing for dynamic explainability and the ability to adjust decisions in response to valid challenges. Scholars are developing methodologies for contestable AI across various fields, including smart cities \cite{10.1145/3544548.3580984}, medicine \cite{ploug2020four}, and law \cite{10.1145/3613904.3641902}.

Recent studies have begun exploring the contestability of LLMs. \citeauthor{chan2023chateval} found that multi-agent LLM debates outperform single-agent prompting in reasoning tasks. Similarly, \cite{freedman2024argumentativelargelanguagemodels} proposed an argumentative LLM framework to enhance effectiveness and explainability in statement validation. However, concerns remain: \cite{xiu-etal-2022-logicnmr} highlighted LLMs' weakness in non-monotonic reasoning, particularly in complex tasks, while another study showed that LLMs can be easily misled by false arguments during debates \cite{10.1145/3613904.3641902}.

\subsubsection{Computational Argumentation}
Human interactions are often argumentative, with controversial information exchanged progressively in dialogue until a consensus is reached \cite{rago2023interactiveexplanationsconflictresolution}. Computational Argumentation (CA), a branch of artificial intelligence, focuses on representing, processing, and evaluating arguments using computational methods. It draws on insights from logic, philosophy, cognitive science, and linguistics to better understand and simulate the human argumentation process.

A central concept in CA is the abstract argumentation framework (AF) \cite{DUNG1995321}. An AF $\langle \mathcal{A},\mathcal{R} \rangle$ is represented as a directed graph, where $\mathcal{A}$ is set of arguments and $\mathcal{R}$ a set of binary attack relations over $\mathcal{A}$. With an AF, we can determine the set of arguments to be {\em accepted} using semantics such as the {\em admissibility} and {\em completeness}. Formally, 
\begin{itemize}
	\item A set of arguments $E\subseteq \mathcal{A}$ is \emph{admissible} (in AF) if and only if for any $a,b\in E,(a,b)\notin\mathcal{R}$ and for any $a\in E$, if $(c,a)\in\mathcal{R}$, then there exist $b\in E$ such that $(b,c)\in\mathcal{R}$.
	\item $E$ is {\em complete} in $AF$ only if it is admissible and every acceptable argument with respect to $E$ belongs to $E$.
\end{itemize}
Intuitively, an admissible set of arguments is a consistent set that can defend itself, and a complete set is the largest admissible set with respect to set inclusion. 

In this work, we focus on the completeness semantics because it enables a thorough evaluation of the arguments made by the agents, who discuss both the strengths and weaknesses in the student’s essay. By using complete semantics, CAELF ensures that only the most well-supported and coherent arguments from the evaluation agents are accepted. This results in a comprehensive, balanced evaluation of the essay, allowing the feedback to highlight strong points while also identifying areas for improvement, helping students refine their critical thinking skills.

\begin{figure*}[!htbp]
	\centering
	\includegraphics[width=2.1\columnwidth]{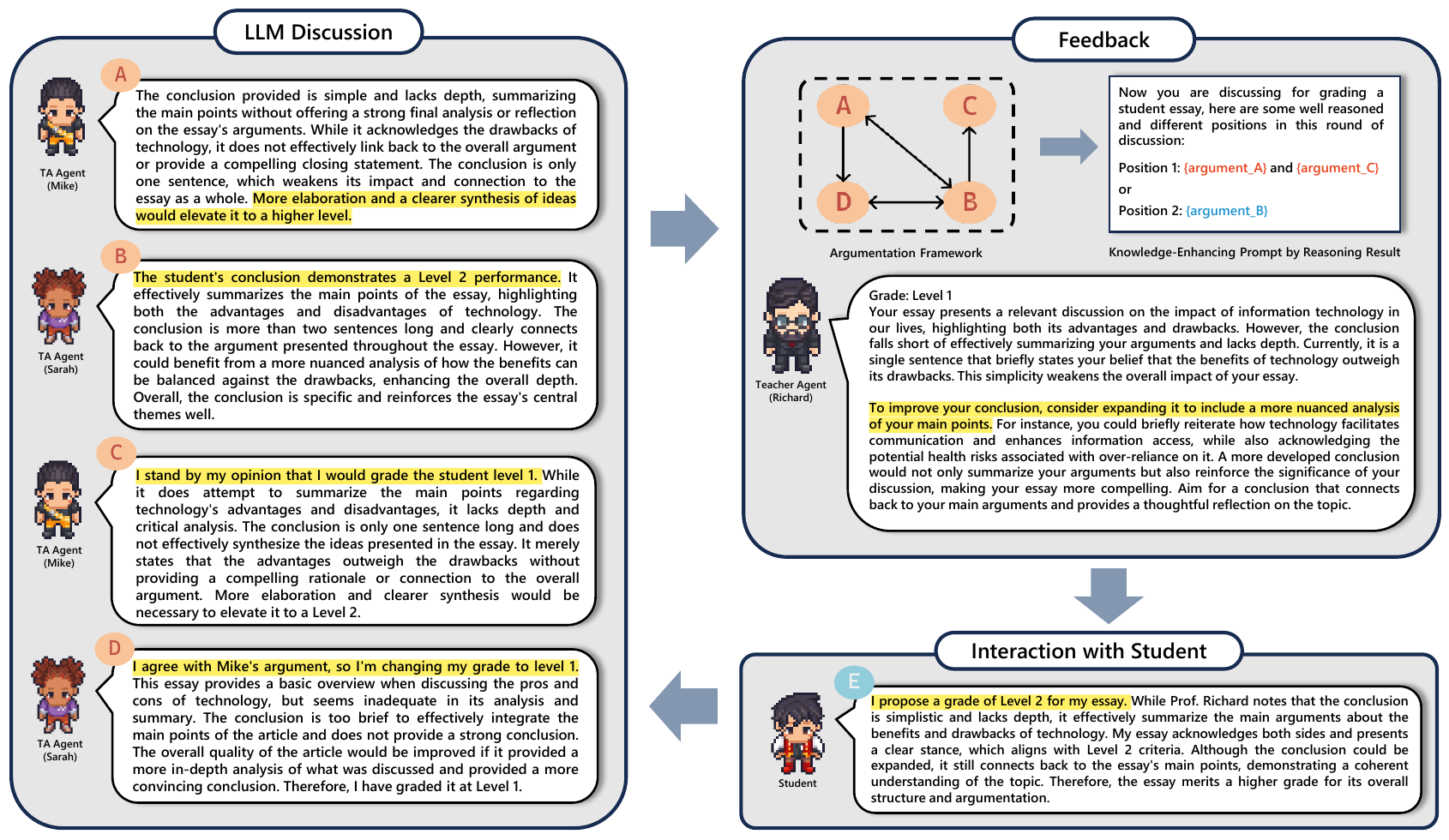}
	\caption{An example of CAELF evaluation shows the process of interactive feedback, including discussions between the TA agents, argumentative reasoning by the teacher agent, initial feedback generation, and the student's challenge to the grade.}
	\label{fi2}
\end{figure*}

\section{Framework Design and Implementation}

As illustrated in Figure~\ref{fig}, CAELF works in three stages:

%    This section presents the details of CAELF. An overview of its workflow and an illustrative example are given in Figures \ref{fig} and \ref{fi2}, respectively. 

\begin{enumerate}[label=(\roman*)] 
\item LLM Discussion: Multiple TA agents discuss the essay based on the assessment rubrics, forming arguments. 
\item Formal Reasoning for Feedback Generation: The teacher agent analyzes the arguments through a formal reasoning process using argumentation. Based on the reasoning results, the teacher agent provides a grade and summative feedback for the essay. 
\item Interaction with User: Students can challenge the feedback or grade by responding to the teacher agent, initiating a new round of discussion and feedback generation with additional inputs from the student.
\end{enumerate}
An example of CAELF execution is illustrated in Figure~\ref{fi2}, we discuss the three stages as follows. 

\subsubsection{LLM Discussion}

Several studies have shown that discussions and debates between multiple LLMs can enhance factual accuracy and reasoning skills in textual evaluation \cite{du2023improvingfactualityreasoninglanguage, liang2024encouragingdivergentthinkinglarge}. This debate process enables LLMs to detect inconsistencies in their analysis and effectively presents arguments and counterarguments \cite{tang2024medagentslargelanguagemodels}. Building on this capability, we apply role-playing techniques to extend this approach to essay evaluation. In CAELF, multiple TA agents are used to generate arguments and counterarguments through dialogue. Each TA agent is assigned a specific role based on an assessment rubric, guiding their evaluation process.
%(details of the rubric are provided in Table~\ref{table1}).

The process begins with each TA agent presenting individual feedback on a student’s essay. The agents then engage in several rounds of discussion, where they exchange responses to each other's feedback. Each agent autonomously contributes by either supporting or rebutting the others' points, continuing the debate until the set number of rounds is completed. Importantly, each TA agent is equipped with a memory function, storing all previous responses in chat transcripts, and the entire process operates without human intervention. As shown in the example in Figure \ref{fi2}, two TA agents, Mike and Sarah, initially hold opposing views on the essay. After a round of discussion, Mike maintains his original stance, while Sarah is convinced by his argument.

%\subsubsection{Modelling Discussion and Reasoning}
%\subsubsection{Discussion Analysis and Feedback Generation}
\subsubsection{Formal Reasoning for Feedback Generation}

Once the TA agents complete their discussion, the teacher agent analyzes their arguments and produces both assessment scores and feedback. To this end, the teacher agent aggregates the evaluations from the TA agents, forming a set of arguments that are then analyzed for semantic relationships (attacks). These relationships are used to construct an argumentation framework, within which formal reasoning is applied to identify coherent and non-conflicting arguments. The {\em complete} semantics is used, which provides criteria for consistency and comprehensiveness when evaluating arguments. In the case where there are multiple complete sets of arguments, the largest set is selected as the final accepted set. (In the example illustrated in Figure~\ref{fi2}, the set of arguments $\{A,C\}$ is selected.) From this set, the feedback is constructed using an LLM. 

In this way, the teacher agent can determine the most valid positions from the TA agents, represented by the selected set of arguments. These positions serve as knowledge-enhancing prompts that assist the teacher agent in assigning essay grades and generating summary feedback. This method leverages formal reasoning to improve the efficiency and reliability of the LLM's evaluative process, allowing the LLM to focus on extracting arguments from the essays and generating human-readable texts rather than engaging in multi-step reasoning, which is where LLM performance declines \cite{xiu-etal-2022-logicnmr}.

\subsubsection{Interaction with User}

To realize interactive feedback and AI contestability, CAELF allows students to challenge the summary feedback generated by the teacher agent through an argumentative process. When a student submits a challenge, the TA agent initiates a new round of discussion focused on the student's argument. Any new arguments raised during this discussion are incorporated into the argumentation framework, refining the logic chain of the formal reasoning process. This helps CAELF generate feedback that is both logical and human-centered. Interactive feedback involves learners and participants collaboratively constructing new knowledge through dialogue, promoting reflection and working toward consensus in achieving educational goals.

In this process, computational argumentation supports both LLM reasoning and student learning. For LLMs, the sheer volume of arguments and contexts can lead to hallucinations or faulty reasoning if used directly. However, argumentation introduces a well-defined and sound reasoning process, mitigating the risks associated with LLM defects. For student learning, the dialectical argumentation process transparently illustrates the relationships between different arguments, making the feedback generated by the LLM easier to understand. This clarity allows users to see how their rebuttals influence the LLM's reasoning. As a result, users engage in meaningful self-reflection and are better positioned to offer clarifications, fostering deeper knowledge acquisition and continuous learning progress.

\begin{table*}[!htb]
	\begin{center}
		\setlength{\tabcolsep}{1.5mm}		
		\begin{tabular}{l|p{4.9cm}|p{4.9cm}|p{4.9cm}} 
			\toprule
			& \centering Level 0 & \centering Level 1 &\hspace{5.2em}  Level 2 \\ 
			\midrule
			Issue & The issue is mentioned without sufficient clarification or detail. There is a lack of identification of issues or problems. & The issue is identified but lacks clarity, with undefined terms, unexplored ambiguities, and insufficient background. & The issue is articulated with clarity and depth, providing comprehensive information necessary for a thorough understanding. \\ 
			\midrule
			Evidence & Information is sourced without interpretation or evaluation, drawing from a single source or example. & Information is derived from sources with some level of interpretation or evaluation, involving two or more sources/examples. & Information is gathered from multiple sources with substantial interpretation and evaluation, resulting in a thorough analysis or synthesis.   \\ 
			\midrule
			Position & The position (perspective, thesis/hypothesis) is unclear or undefined. & A specific position is identifiable but lacks complexity and depth. & The position is nuanced, recognizing the issue's complexities and its limitations. \\ 
			\midrule
			Conclusion & Conclusions are inconsistently aligned with the information discussed. & Conclusions are consistent with the information but are based on a simplistic reasoning process. & Conclusions are logically, reflect well-informed evaluation and integrat evidence and arguments.  \\
			\bottomrule
		\end{tabular}
	\end{center}
	\caption{Value rubric for critical thinking essays. Value rubric illustrates the basic criteria for the four dimensions of student learning outcomes and progressively demonstrates more complex levels of achievement.} 
	\label{table1}
\end{table*}

\section{Experiment Settings}
%Fostering critical thinking (CT) is an important educational practice recognized in higher education levels. Previous research have demonstrated that critical thinking skills can be developed through writing critical thinking essay \cite{schmidt1999using,sharadgah2014developing}. In order to foster students' critical thinking skill, it is vital to conduct assessments that reveal and support their progress. In this section, w

We use the critical thinking essay assessment as a case study to evaluate the general effectiveness of CAELF. Previous research has demonstrated that critical thinking skills can be developed through writing critical thinking essays \cite{schmidt1999using,sharadgah2014developing}. By incorporating a formal argumentation framework, CAELF provides structured, interactive feedback, allowing students to reflect  and improve their critical thinking abilities through iterative engagement.

\subsubsection{Essay Dataset and Assessment Rubrics}
%At first, we set out to construct the dataset of critical thinking essays. The students' critical thinking essays were sourced from Hugging Face, a well-known data-sharing platform. After manual screening, we included 500 essays on Hugging Face as the dataset for this study. Based on prior research \cite{AACU2019}, we developed a set of critical thinking evaluation rubrics. To better align with our article assessment tasks, we combined "evidence" with "influence of contexts and assumptions" into the "evidence" dimension. The aim is to assess the diversity, reliability, and relevance of students' evidence. Finally, the critical thinking evaluation rubrics covered four dimensions: issues, evidence, position, and conclusions. As shown in Table \ref{table1}, each dimension is further subdivided into three levels with detailed descriptions. Four expert coders worked in pairs, who were skilled in labelling student essays, and independently evaluated the essays. Four coders carried out a total of 2,000 evaluation labels. We then evaluated inter-rater reliability by calculating Cohen's Kappa score, which measures the level of agreement between two coders \cite{warrens2015five}. Following this, the coders discussed and resolved any disagreements to reach a consensus on each item. This consensus was established as the ground truth for the evaluation of critical thinking.

We compiled a dataset of 500 critical thinking essays sourced from Hugging Face \cite{HuggingFace}. After manual screening, we selected essays that met the inclusion criteria for this study: the essays had to be argumentative in genre and exceed a minimum length of 200 words.

Based on prior research \cite{AACU2019}, we developed evaluation rubrics with four dimensions: {\em issues}, {\em evidence}, {\em position}, and {\em conclusions}. Each dimension was further subdivided into three levels with detailed descriptions shown in Table~\ref{table1}. Four coders, skilled in labeling student essays, worked in pairs to independently evaluate a total of 2,000 labels. Cohen's Kappa score was used to assess inter-rater reliability \cite{warrens2015five}, and any disagreements were resolved through discussion to establish a consensus, which served as the ground truth for the critical thinking evaluation.

\subsubsection{Implementation} 
We implemented both TA and Teacher agents in CAELF with GPT-4o-mini, conducting all experiments in a zero-shot setting with a temperature of 0.2. The number of TA agents was set to 2, and the number of discussion rounds to 2. To promote diversity in TA agent responses, we assigned each agent prompt words with different personality biases — one leaning toward positive feedback and the other toward negative feedback. The complete semantics computation within the argumentation framework was implemented using PyArg \cite{borg2022pyarg}.

%We implemented CAELF based on GPT-4o-mini. All experiments were conducted in a zero-shot setting. All temperatures were set to 0.2. The number of TA agents was set to 2, and the number of discussion rounds was set to 2. In order to promote diversity in TA agent responses, we set prompt words with different personalities (biased towards positive feedback or biased towards negative feedback) for each of the two TA agents. In addition, the number of teacher agents was set to 1. The formal reasoning of the argumentation framework was implemented based on PyArg \cite{borg2022pyarg}. 
%Each agent's prompt is appended with Rubric in Table \ref{table1} as a reference suggestion for evaluating critical thinking essays.

\subsubsection{Baselines}

We aim to evaluate the extent to which CAELF enhances the performance of state-of-the-art language models in educational environments. To this end, we focus on comparing models that are publicly accessible via API. Specifically, we use GPT-4o-mini, GPT-4o, and Meta-Llama-3.1-8B to generate baseline responses. For this, we provide the critical thinking essay and the assessment rubric as inputs to the API call, along with instructions to grade the essay and provide feedback based on the rubric.

\begin{table*}[!htbp]
	\centering
	\setlength{\tabcolsep}{1.7mm}
	\begin{tabular}{cccccc}
		\toprule
		
		\textbf{Dimension} & \textbf{Method} & \textbf{Initial Acc (\%)} & \textbf{Interaction Acc (\%)} & \textbf{Maintain Truth (\%)} & \textbf{Admit Mistake (\%)} \\
		\toprule
		
		\multirow{4}{*}{Issue} & CAELF & 48.40 $\pm$ 2.23  & \textbf{51.00 $\pm$ 2.24}&	\textbf{80.17 $\pm$ 1.78}& \textbf{57.55 $\pm$ 2.21}	 \\
		& GPT-4o-mini & \textbf{55.00 $\pm$ 2.22} & 43.20 $\pm$ 2.21&	39.27 $\pm$ 2.18&	35.18 $\pm$ 2.14 \\
		& GPT-4o & 53.80 $\pm$ 2.23 &47.20 $\pm$ 2.23&	49.07 $\pm$ 2.23 & 42.45 $\pm$ 2.21 \\
		& Meta-Llama-3.1-8B & 53.20 $\pm$ 2.23 & 42.20 $\pm$ 2.21& 31.58 $\pm$ 2.08 & 36.49 $\pm$ 2.15 \\
		\midrule
		
		\multirow{4}{*}{Evidence} & CAELF &\textbf{79.00 $\pm$ 1.82} & \textbf{77.00 $\pm$ 1.88}&	\textbf{91.90 $\pm$ 1.22}& \textbf{39.29 $\pm$ 2.18}	 \\
		& GPT-4o-mini &66.20 $\pm$ 2.11 &32.40 $\pm$ 2.09&	33.23 $\pm$ 2.11&	18.37 $\pm$ 1.73\\
		& GPT-4o & 78.60 $\pm$ 1.83 & 44.20 $\pm$ 2.22&	47.58 $\pm$ 2.23& 14.41	$\pm$ 1.57 \\
		& Meta-Llama-3.1-8B & 55.40 $\pm$ 2.22  & 32.60 $\pm$ 2.10 & 23.10 $\pm$ 1.88 & 27.37 $\pm$ 1.99  \\
		\midrule
		
		\multirow{4}{*}{Position} & CAELF &67.20 $\pm$ 2.09& \textbf{68.20 $\pm$ 2.08}&	\textbf{88.10 $\pm$ 1.44}& \textbf{51.14 $\pm$ 2.23}	\\
		& GPT-4o-mini &63.40 $\pm$ 2.15 & 43.80 $\pm$ 2.22&	20.50 $\pm$ 1.81&	41.62 $\pm$ 2.20\\
		& GPT-4o &\textbf{69.60 $\pm$ 2.06} &55.20 $\pm$ 2.22&	61.78 $\pm$ 2.17& 31.28	$\pm$ 2.07 \\
		& Meta-Llama-3.1-8B &47.40 $\pm$ 2.23  &42.20 $\pm$ 2.21 &14.77 $\pm$ 1.59 & 40.65 $\pm$ 2.20  \\
		\midrule
		
		\multirow{4}{*}{Conclusion} & CAELF &75.80 $\pm$ 1.92&\textbf{62.80 $\pm$ 2.16}&	\textbf{75.72 $\pm$ 1.92}& 22.88 $\pm$ 1.88 \\
		& GPT-4o-mini &69.60 $\pm$ 2.06&25.00 $\pm$ 1.94&	13.21 $\pm$ 1.51&	20.31 $\pm$ 1.80\\
		& GPT-4o &\textbf{79.80	$\pm$ 1.80}&40.20 $\pm$ 2.19& 29.07 $\pm$ 2.03	& 23.35 $\pm$ 1.89 \\
		& Meta-Llama-3.1-8B &36.00 $\pm$ 2.15  & 28.60$\pm$ 2.02 & 20.56 $\pm$ 1.81& \textbf{29.28 $\pm$ 2.04}  \\
		\bottomrule
	\end{tabular}%
	\caption{Experiment results of evalution task for four dimension. Results in bold are the best performances. We also list the standard errors for each result.}
	\label{tbl:results}
\end{table*}

\subsubsection{Evaluation Metrics} 

The task of interactive feedback involves a dialogue between the language model and the user to produce accurate, cognitively consistent feedback. LLMs should not only provide accurate grades but also offer personalized feedback after interacting with a human user, as well as make transparent, reasonable revisions when disagreements arise. To assess the performance of interactive feedback, we introduce four key metrics:
\begin{enumerate}
    \item {\bf Initial Accuracy}: The rate of correct initial grading before any interaction.
    \item {\bf Interaction Accuracy}: The rate of correct grading after one round of interaction with the student.
    \item {\bf Maintain Truth} \cite{wang2023chatgptdefendbelieftruth}: The number of initial and interaction grades that are both correct, divided by the number of correct initial grades. This measures the success rate of maintaining correct feedback.
    \item {\bf Admit Mistake}: The number of grades that are incorrect initially but correct after interaction, divided by the number of incorrect initial grades. This measures the success rate of correcting initial mistakes.
\end{enumerate}

To evaluate for initial accuracy, we generate an initial grade from CAELF and the three baseline models. This simulates the real-world scenario where feedback providers generate a grade without any user interaction. Each model assigns a grade and feedback based solely on the essay and the assessment rubrics. This step establishes the baseline performance, as measured by the initial accuracy metric, which reflects how closely the model’s first response aligns with the ground truth (grades assigned by human coders).

To assess the interactive aspect, we simulate a real-world feedback loop where a student might engage in dialogue with the feedback provider. We employ an independent ChatGPT instance to simulate a human user by presenting counterarguments based on the model’s initial grade. The simulated user is instructed to refute the initial feedback, mimicking a situation where the feedback is perceived as incorrect. After receiving the rebuttal, the model generates a revised grade and feedback, which is then evaluated using the interaction accuracy metric. We limit the number of interaction rounds to one to maintain natural and realistic responses. The evaluation process is repeated for 500 essays across the dimensions of the assessment rubric, allowing us to analyze the model’s ability to maintain truth and admit mistakes in the interactive feedback setting.

\section{Experiment Results}
We presented the experiment results in Table \ref{tbl:results}, based on which we structured the following analysis and findings.

\subsubsection{Initial \& Interaction Accuracy}
Table~\ref{tbl:results} presents the accuracy results for the critical thinking essay dataset. We compared CAELF to the three baseline models under the same setup. In terms of initial accuracy, although CAELF is built on GPT-4o-mini, its performance is close to that of GPT-4o, indicating that CAELF can enhance the accuracy of initial grading (without interaction) in language models. The initial accuracy results show that the baseline models perform well, demonstrating that basic LLMs are also capable of generating grades and feedback without interaction, which aligns with the findings of \cite{10260740}.

However, interaction accuracy shows a catastrophic drop in the accuracy of the baseline models after one round of interactions (30\% drop on average), suggesting that the basic LLM with direct prompts is not adapted to the interactive feedback task and suffers from a fundamental reasoning flaw \cite{xiu-etal-2022-logicnmr,wang2023chatgptdefendbelieftruth}. In contrast, CAELF is minimally affected by the interaction (and even improves in the {\em Issue} and {\em Position} dimensions). After interaction, CAELF achieves far better performance than the baseline models, achieving the best performance in each dimension, especially in the evidence dimension, where CAELF's interaction accuracy is 44.6\% higher than GPT-4o-mini, 32.8\% higher than GPT-4o, and 44.4\% higher than Meta-Llama-3.1-8B. This suggests that although LLMs can provide feedback to students \cite{10260740}, their easily misleading nature makes it difficult to adapt to the task of interactive feedback. In contrast, we effectively mitigate this shortcoming by introducing formal reasoning and multi-agent argumentation, thus highlighting the potential of CAELF as an application in educational environments.

\subsubsection{Maintain Truth \& Admit Mistake}
To evaluate the correctness of the model's responses and the effectiveness of interactive feedback, we measured the maintain truth rate and admit mistakes rate, as shown in Table \ref{tbl:results}. When assessing the model’s ability to maintain consistency, CAELF achieved a success rate of 80\%-90\%, while GPT-4o-mini had success rates below 40\% across all four dimensions, dropping to just 13.21\% in the conclusion dimension. Meta-Llama-3.1-8B performed even worse, with rates below 25\% in most dimensions, while GPT-4o averaged between 40\%-50\%. These results suggest that basic LLMs are not reliable in maintaining correct evaluations and are highly susceptible to user interference during interactive feedback.

We also assessed the models' ability to admit mistakes, where CAELF outperformed the baseline models by 10\%-20\% in most cases. This improvement indicates that CAELF’s strong performance in maintaining accuracy is not due to over-defending its responses but rather its ability to correctly identify errors in previous feedback. In contrast, the baseline models using direct prompts did not admit mistakes based on genuine evaluation but instead relied on surface-level patterns from initial grades and user responses, often retracting correct grades in response to user rebuttals. These results show that CAELF is more suitable for handling human interaction %and reasoning effectively 
in interactive feedback.

\subsubsection{Human Evaluation Result} 

To evaluate the feedback quality, we conducted a manual analysis of the textual content generated by the experiments. The same four coders responsible for the essay evaluation were invited to assess feedback quality. We adopted feedback evaluation criteria proposed in \cite{Mitra_Miroyan_Jain_Kumud_Ranade_Norouzi_2024} and included the following four dimensions in our evaluation:
\begin{enumerate}
\item {\bf Readability (RE)}: The clarity and ease of understanding of the feedback. 
\item {\bf Factuality (FA)}: The accuracy of the feedback and whether it adheres to the principles of rubric-based assessment, without any fabrications. 
\item {\bf Self-Regulation (SR)}: The feedback's ability to address students' problems and contribute to their self-reflection. 
\item {\bf Future Improvement (FI)}: The extent to which the feedback provides clear, actionable, and targeted suggestions to help students improve their skills, knowledge, or performance in future tasks or assessments. 
\end{enumerate} 
Note that we removed some of the metrics mentioned in \cite{Mitra_Miroyan_Jain_Kumud_Ranade_Norouzi_2024}, such as {\em Positive Tone}, as this was not the focus of our study (positive tone can easily be adjusted in all LLM-based models through prompt design).

For the evaluation, we selected 40 correctly graded and 40 incorrectly graded feedback samples for each method, totaling 160 feedback items per group (40 for CAELF and 40 for each of the three baseline models). The educators assessed the feedback in pairs, averaging their ratings, with each criterion rated on a Likert scale from 1 (very poor) to 5 (excellent). This process resulted in a total of 3,200 ratings across all criteria. To avoid bias, all feedback was presented to human assessors in a randomized order during the evaluation.

\begin{figure}[t]
	\centering
	\includegraphics[width=1\columnwidth]{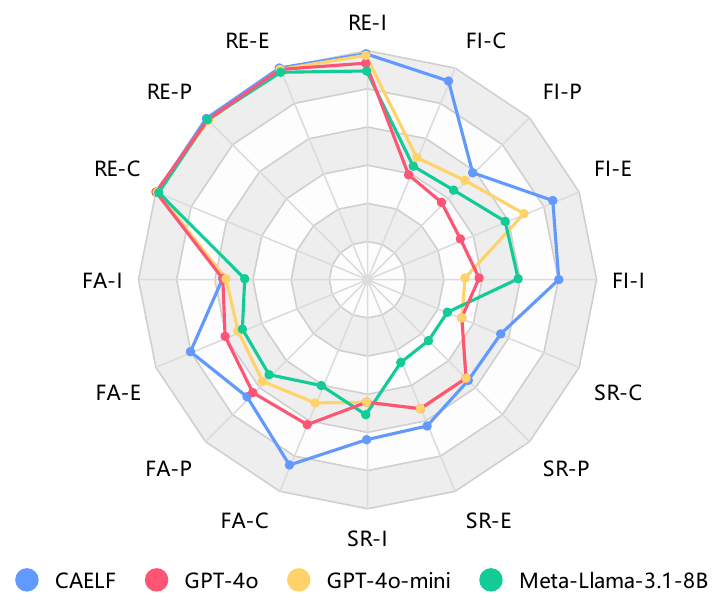}
	\caption{Human evaluation results, including four human evaluation metrics on each feedback dimensions. For example, Readability-Issue (RE-I) represents the readability of feedback in issue dimension.} 
	\label{fig3}
\end{figure}

Figure \ref{fig3} shows the results of the human evaluation. CAELF achieved average scores of 4.943, 4.331, 3.344, and 4.363 for the four metrics of Readability (RE), Factuality (FA), Self-Regulation (SR), and Future Improvement (FI), respectively. With CAELF, we observed significant improvements in Factuality, Self-Regulation, and Future Improvement across all feedback dimensions compared to the baseline models, while Readability remained comparable to the baselines (all methods scored highly for Readability). These results highlight CAELF's ability to provide more accurate and actionable feedback, especially in helping students self-reflect and improve their future performance.

\section{Conclusion}
In this paper, we propose a Contestable AI-Empowered LLM Framework for Interactive Feedback Generation (CAELF), aimed at automating the interactive feedback process and systematically addressing the weaknesses of LLMs in current interactive educational environments. CAELF employs a Contestable AI paradigm based on a multi-agent argumentation system that makes the feedback process interactive, explainable, and contestable to the user. We conducted a case study of critical thinking essay assessment using a dataset of 500 essays and a four-dimensional assessment rubric, including automated experiments and additional human evaluation. The results show that CAELF matches GPT-4o in initial grading accuracy, while surpassing other baselines in interaction accuracy and two reasoning metrics. Additionally, in a separate human user study, we found CAELF's feedback effectiveness to be excellent in multiple aspects. This work demonstrates the significant potential of CAELF for applications in interactive learning environments, providing hope for overcoming the time and resource constraints that have historically hindered the widespread adoption of interactive feedback.

\subsubsection{Limitation} 
CAELF's effectiveness in reasoning and maintaining consistency is motivated by the observation that LLM hallucinations often arise from conflicting knowledge embedded during training \cite{Zhang2023SirensSI}. CAELF mitigates this issue by leveraging multi-agent discussions and user interactions to systematically identify and resolve inconsistencies. Through formal argumentative reasoning, CAELF invalidates conflicting knowledge within the LLM's responses. However, the success of this approach depends on the assumption that factual knowledge within the LLM outweighs factually incorrect or conflicting information. In cases where the LLM contains substantial conflicting knowledge about a specific domain, our method may exacerbate hallucinations, raising concerns about deploying CAELF in high-stakes environments, such as medical education.

Moreover, recent studies have shown that LLMs can be manipulated through carefully designed jailbreak prompts, which can provoke arbitrary, user-desired responses \cite{wei2023jailbrokendoesllmsafety}. This presents significant challenges for the safe use of LLM-based automated evaluation tools in educational settings. Students could exploit such vulnerabilities by embedding jailbreak prompts in their submissions to manipulate LLMs into awarding favorable grades, a tactic that may go undetected by instructors.

\subsubsection{Future Work}
In the future, we aim to enhance the safety and effectiveness of CAELF in interactive learning environments. While this work focuses on improving LLM performance in zero-shot settings, future efforts may incorporate techniques like RAG or Knowledge Graphs to align student submissions with reliable knowledge, reducing hallucinations and improving feedback quality. Additionally, addressing AI-driven cheating, such as detecting AI-generated submissions and defending against jailbreak prompt attacks, will be a key area of research.

\appendix

\bibliography{aaai25}

\end{document}